\documentclass{article}

\usepackage{arxiv}

\usepackage[utf8]{inputenc} % allow utf-8 input
\usepackage[T1]{fontenc}    % use 8-bit T1 fonts
\usepackage{hyperref}       % hyperlinks
\usepackage{url}            % simple URL typesetting
\usepackage{booktabs}       % professional-quality tables
\usepackage{amsfonts}       % blackboard math symbols
\usepackage{nicefrac}       % compact symbols for 1/2, etc.
\usepackage{microtype}      % microtypography
\usepackage{lipsum}
\usepackage{graphicx}
\graphicspath{ {./images/} }

\usepackage{hyperref}
\makeatletter
\newcommand{\printfnsymbol}[1]{%
  \textsuperscript{\@fnsymbol{#1}}%
}
\makeatother
\usepackage{amsmath}
\usepackage{multirow}
\usepackage{makecell}
\usepackage[numbers]{natbib}

\title{Theme-Matters: Fashion Compatibility Learning via Theme Attention}
\author{Jui-Hsin(Larry) Lai\textsuperscript{\rm 1}\thanks{This work is completed during the time in Silicon Valley Research Center, JD.com. Jui-Hsin(Larry) Lai and Bo Wu have equal contribution.}, Bo Wu\textsuperscript{\rm 3}\printfnsymbol{1}, Xin Wang\textsuperscript{\rm 4}, Dan Zeng\textsuperscript{\rm 5}, Tao Mei\textsuperscript{\rm 2}, Jingen Liu\textsuperscript{\rm 2} \\
\textsuperscript{\rm 1}PAII-Labs, \textsuperscript{\rm 3}Columbia University, \textsuperscript{\rm 2}JD.com, \textsuperscript{\rm 4}Donghua University, \textsuperscript{\rm 5}Shanghai University \\ %If you have multiple authors and 
juihsin.lai@gmail.com, bo.wu@columbia.edu, wangx@mail.dhu.edu.cn, \\dzeng@shu.edu.cn, tmei@live.com, jingen.liu@gmail.com
}

\begin{document}
\maketitle
\begin{abstract}
Fashion compatibility learning is important to many fashion markets such as outfit composition and online fashion recommendation. Unlike previous work, we argue that fashion compatibility is not only a visual appearance compatible problem but also a theme-matters problem. An outfit, which consists of a set of fashion items (e.g., shirt, suit, shoes, etc.), is considered to be compatible for a ``dating'' event, yet maybe not for a ``business'' occasion. In this paper, we aim at solving the fashion compatibility problem given specific themes. To this end, we built the first real-world theme-aware fashion dataset comprising 14K around outfits labeled with 32 themes. In this dataset, there are more than 40K fashion items labeled with 152 fine-grained categories. We also propose an attention model learning fashion compatibility given a specific theme. It starts with a category-specific subspace learning, which projects compatible outfit items in certain categories to be close in the subspace. Thanks to strong connections between fashion themes and categories, we then build a theme-attention model over the category-specific embedding space. This model associates themes with the pairwise compatibility with attention, and thus compute the outfit-wise compatibility. To the best of our knowledge, this is the first attempt to estimate outfit compatibility conditional on a theme. We conduct extensive qualitative and quantitative experiments on our new dataset. Our method outperforms the state-of-the-art approaches. 
\end{abstract}

% keywords can be removed
%\keywords{Fashion Compatibility, Fashion Outfit, Metric Learning, Deep Embedding}

\section{Introduction}
Fashion compatibility learning, whose goal is to automatically compose compatible outfits for recommendations, is of importance to a variety of academic and industrial tasks such as outfit composition~\cite{feng2018interpretable}, wardrobe creation~\cite{hsiao2018creating}, item recommendation~\cite{he2016learning}, fashion generation~\cite{bettaney2019fashion}. It has recently attracted increasing attention ~\cite{han2017learning,Shih2017,li2017mining,Tangseng2018,Nakamura2018,Vasileva2018,hsiao2017learning,simo2016fashion,he2016learning,veit2015learning,song2017neurostylist}. 

\begin{figure}[t]
	\begin{center} 
		\includegraphics[width=0.7\linewidth]{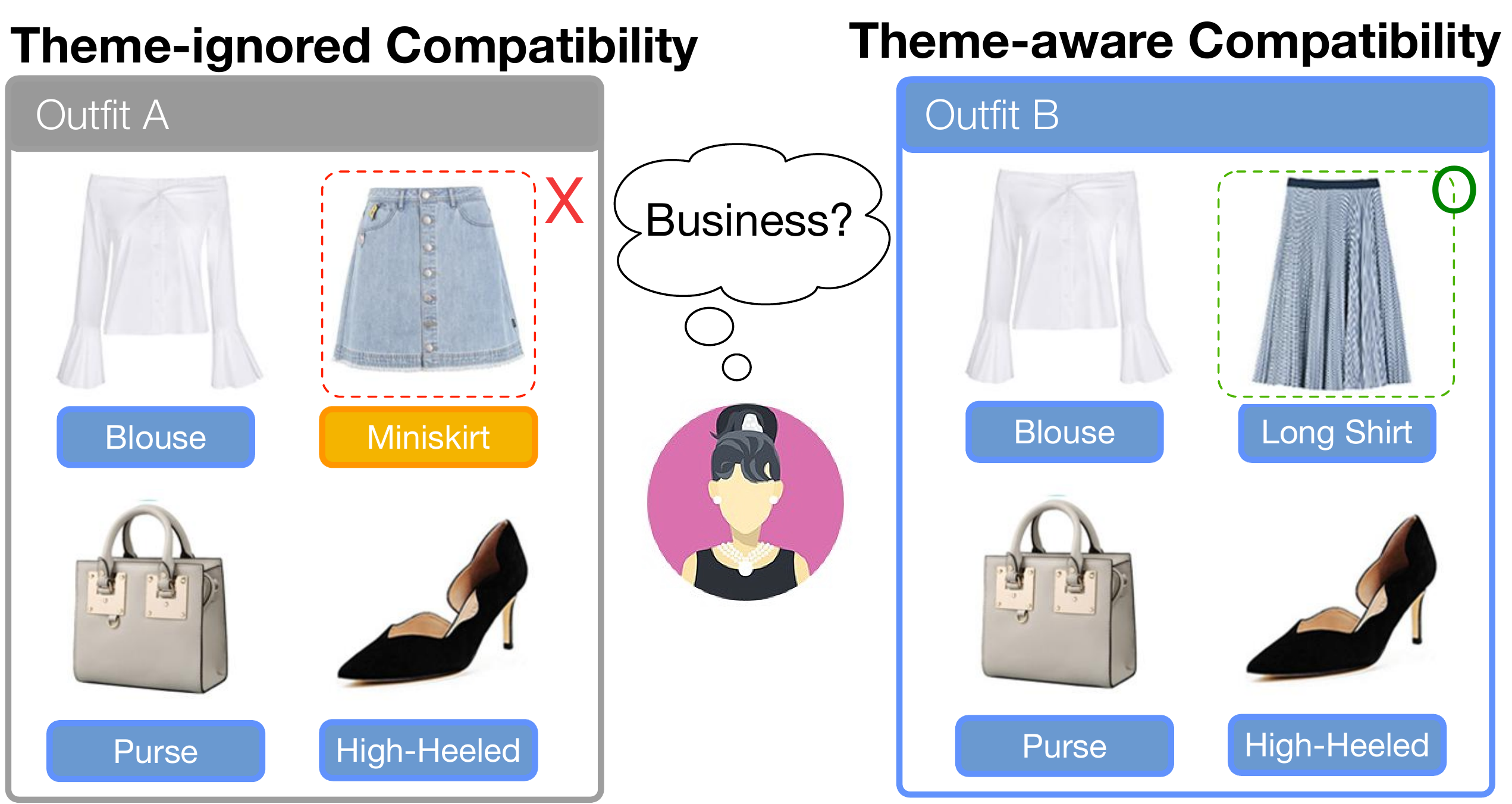} 
	\end{center}   
  
	\caption{An example demonstrating the importance of theme-aware fashion compatibility. In the theme-ignored case, Outfit A may be compatible based on visual appearance. But, ``is it suitable for the business occasion?'' In general, miniskirt may not fit an office. So in the theme-aware case, we can generate Outfit B, which may be better for business.} 
	%\caption{Examples for outfit difference of fashion outfits with different themes. From the visual perception, the compatibility of compared outfits are similar, but the theme of them are totally different.}
	\vspace{-0.2in}
	\label{fig:task}
\end{figure}

In general, we can categorize the fashion compatibility learning methods into two classes: one formulates it as a pair-wise learning task~\cite{McATarShiHen15}~\cite{Veit2017}~\cite{Shih2017}~\cite{Vasileva2018}, which develops measurement methods (e.g., metric learning ) for pair-wise compatibility, and another one is outfit-wise compatibility learning, which models the process of forming an outfit as sequence learning (i.e., LSTM)~\cite{han2017learning}~\cite{li2017mining}~\cite{Tangseng2018}. Most existing works treat fashion compatibility as visual appearance compatible problem. As a result, although significant progress has been made, we are still not able to answer a question as shown in Figure~\ref{fig:task}: ``is Outfit A compatible in a business occasion?''.

Fashion compatibility is also a theme-matters problem. For example, as shown in Figure~\ref{fig:task}, Outfit A may be compatible based on visual appearance and can be dressed for dating. But if one wants to have it for business, she may want to adjust it to Outfit B (long shirt instead of miniskirt for business). Therefore, theme-aware fashion compatibility is very important for fashion recommendation.

Most existing fashion datasets such as Polyvore dataset~\cite{han2017learning} and DeepFashion2~\cite{DeepFashion2}, however, do not carry the capability to estimate theme-aware fashion compatibility. Hence, we built a new real-world fashion dataset called Fashion32, which is the first one with rich annotations including outfit themes and fine-grained fashion categories. Since the annotations were labeled by fashion stylists from brand vendors, they generally are of high quality. Fashion32 contains 32 theme tags for more than 13K around outfits, and 152 fine-grained categories for more than 40K outfit items. To learn theme-aware fashion compatibility models from this dataset, we face two challenges: how to measure pair-wise compatibility of outfit items and how to associate a theme to pairwise compatibility to compute outfit-wise compatibility.    

\begin{figure*}[t] 
	\begin{center}
		\includegraphics[width=1.0\linewidth]{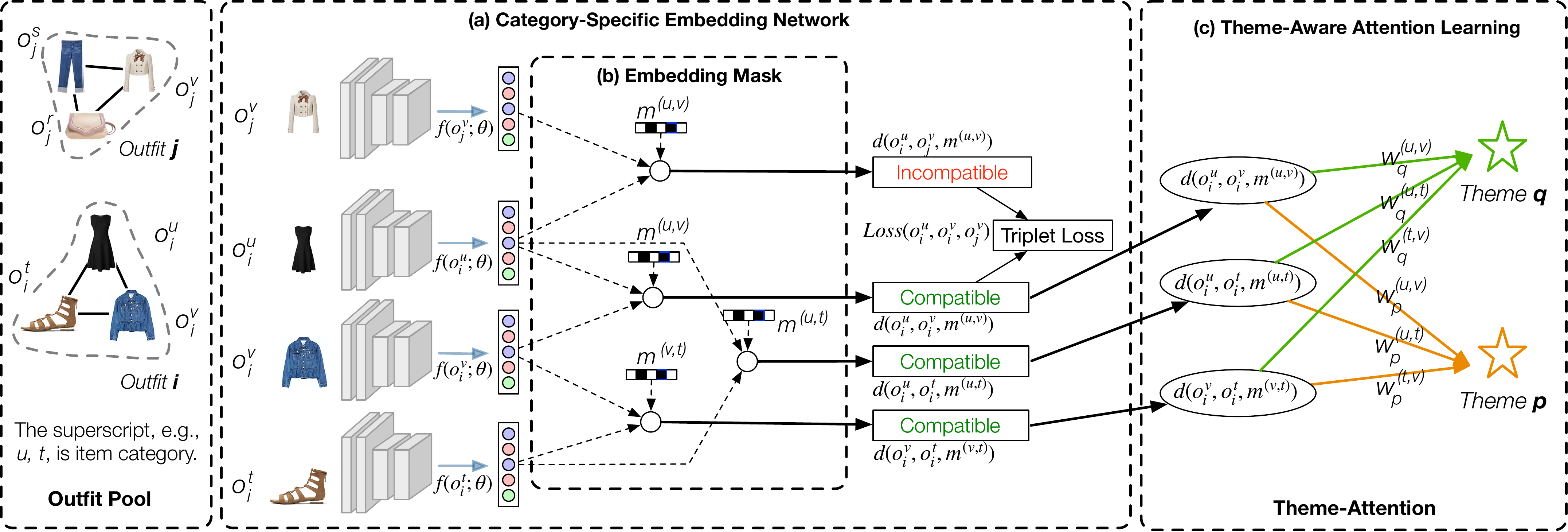} 
	\end{center}  
	\caption{The framework of the proposed theme-attention model for fashion compatibility. This model is built on the fine-grained category, which serves as a bridge connecting outfit/fashion items to the theme. (a) Pairwise outfit items are projected into the category-specific embedding space, and then (c) theme-aware attention is employed to associate themes with the pairwise fashion compatibility. The outfit-wise compatibility is computed as the aggregation of pairwise compatibility using theme-aware attention. }
	\label{fig:framework}
	\vspace{-0.15in} 
\end{figure*}

To address the above challenges, we propose a theme-attention model, which is built on the category-specific embedding space. Figure~\ref{fig:framework} illustrates the overview of our framework. Given an outfit and a specific theme, pair-wise items are projected into the category-specific subspace (Figure~\ref{fig:framework} (a)). Unlike traditional embedding, which maps all fashion items into a common space, we employ triplet network and embedding masks (Figure~\ref{fig:framework} (b)) to project items category-specific subspace embedding. This ``task-orientated'' embedding enables the subspace to be more discriminative for compatibility computing. We further build a theme-attention model to associate the themes with pairwise compatibility (Figure~\ref{fig:framework} (c)). As a result, a theme-specific attention matrix is learned to link the theme to pairwise compatibility of outfit items, and further to aggregate pairwise ones to estimate the outfit-wise compatibility.

%To validate the proposed approach, we implemented compatibility prediction and fill-in-the-blank (FITB) tasks on proposed theme-aware fashion dataset and compared the proposed approach with others on reorganized Polyvore dataset~\cite{han2017learning}.  

%,fashion culture~\cite{chang2017fashion} and fashion trend forecasting~\cite{al2017fashion}. 
%is a practical challenge due to the fact that outfit composition is a process that involves complex interactions of human creativity, fashion expertise and subjective self-expression. 

To the best of our knowledge, our work is the first one to explicitly estimate fashion compatibility given a specific theme. ~\cite{han2017learning} maybe able to answer a question like ``what to dress for a biz meeting'' thanks to their visual-semantic embedding, but their capability relies on the quality of the image captions. Also, their Bi-LSTM framework is less flexible due to specific item order and number. Yet our theme-aware model does not have such constraints. The category-specific embedding is inspired by \cite{Veit2017}\cite{Vasileva2018}. But unlike ~\cite{Veit2017} and ~\cite{Vasileva2018}, which simply group fashion items into coarse categories (e.g., top, bottom, shoe, etc), we employ fine-grained categories because they usually have strong connections to fashion themes due to their properties. For instance, T-shirts imply causal, shirts are more official, and Polo-shirts are in-between. All the properties essentially imply some fashion themes. The coarse category does not carry this advantage.
 
%In addition, fine-grained categories also enable the propose of Theme-Attention due to its strong connections to fashion themes.    
To summarize, our work has the following contributions:

\begin{itemize}

\item We introduce the first theme-aware fashion dataset, which enables to compute the fashion compatibility given a specific theme. The dataset is available to download via the webpage www.larry-lai.com/fashion.html

\item We propose a theme-attention model to associate themes with pairwise compatibility to compute outfit-wise compatibility. To the best of our knowledge, this is the first attempt to study the  theme-matters compatibility learning problem.

\item We leverage fine-grained categories and the category-specific embedding to effectively support our theme-attention model.  

\item We demonstrate our proposed approach can outperform state-of-the-art approaches on Fashion32 dataset and the improved Polyvore dataset \cite{han2017learning}.   
\end{itemize}

\section{Related Work}

\textbf{Fashion Datasets.} In general, we can group the existing fashion datasets, which are built for fashion compatibility learning, into two categories: online shopping datasets~\cite{yamaguchi2015mix,he2016ups,hsiao2018creating} and social media datasets~\cite{han2017learning,Vasileva2018,Tangseng2018}. The former datasets are mined from some online e-commerce platforms by leveraging buyer's shopping carts to form various outfits and labels. As we know, however, a shopping cart usually contains mixed items, which may not form a compatible outfit. Therefore, the labels in online shopping datasets can be very noisy. The social media datasets, such as Maryland Polyvore~\cite{han2017learning} and Polyvore Outfits~\cite{Vasileva2018}, are collected from social media platforms. Since the outfits are created by fashion enthusiasts, these datasets have less outfit noise. But the quality of outfits is very diverse, because outfits created on social media largely depends on users' uploading and labels are also random and bias.

Unlike previous datasets, our Fashion32 not only carries fashion themes, fine-grained categories, and recommendation descriptions but also has high-quality outfit model pictures and abundant annotations. More importantly, our outfit composition and its annotations come from fashion designers of brand vendors. Consequently, our dataset is more realistic and convincing.  

Another two popular fashion datasets, i.e., DeepFashion  ~\cite{liu2016deepfashion} and DeepFashion2~\cite{DeepFashion2}, are specifically built for fashion research including attribute prediction, image retrieval, and fashion synthesis, but not for fashion compatibility.  

%As an example, a shopping cart may contain kid's shoe, mom's dress and dad's sock, which actually is a very common scenario.

\begin{figure}[t]
	\begin{center}
		%\fbox{\rule{0pt}{2in} \rule{1.0\linewidth}{0pt}}
		\includegraphics[width=0.6\linewidth]{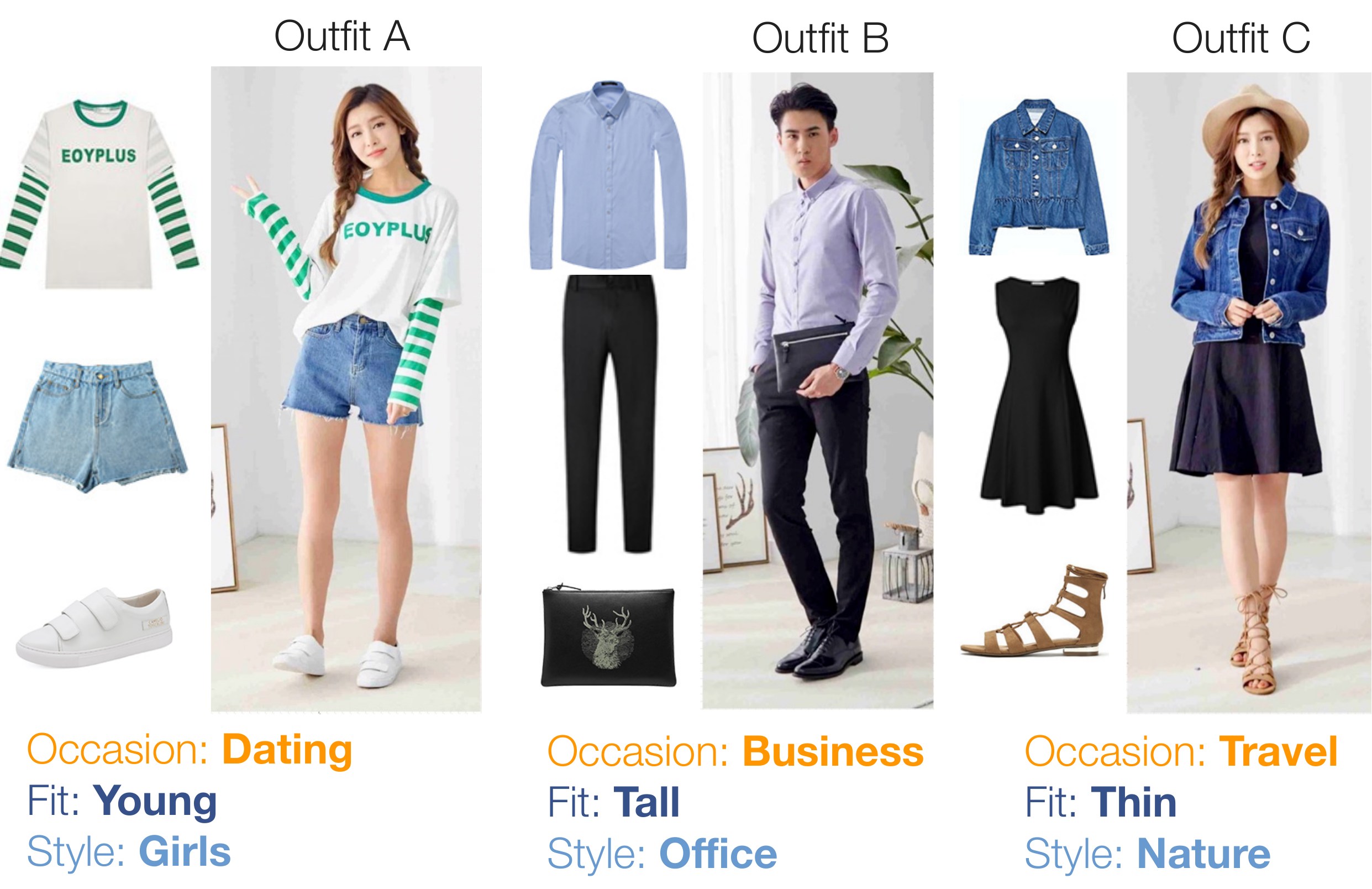}  
	\end{center} 
	\caption{Sample outfits from the Fashion32 dataset. Each outfit carries theme tags, descriptions, items, and pictures of a model to illustrate its compatibility. 
	%b) Polyvore dataset: There is no fine-grained categories, theme information or model photos in previous dataset. 
	}
	\label{fig:FashionTheme1}
	\vspace{-0.2in}
\end{figure}

\textbf{Fashion Compatibility Learning.} In general, telling an outfit compatible or not is a subjective task. One needs to check all possible compatible relationships between items, which can involve very subtle difference. 
One current solution is to leverage metric learning~\cite{McATarShiHen15,Vasileva2018,chen2018dress} or embedding techniques~\cite{van2012stochastic,veit2015learning,wilber2015learning} to project the fashion items into a specific space, in which the outfit compatibility is explicitly measured by pairwise items' distances~\cite{Veit2017}. This measurement is built on pairwise compatibility rather than outfit-wise (namely, computing outfit compatibility as a whole). Han et al~\cite{han2017learning} employ Bi-LSTM beyond visual-semantic embedding to estimate outfit compatibility in an end-to-end model. Meanwhile,Vasileva et al~\cite{Vasileva2018} proposes a category-aware embedding approach to include garment/item types or coarse-grained categories (e.g., top, bottom, etc.) during learning. 
Taking the garment type into consideration, the embedding space consists of a set of type-specific sub-spaces, which further improves the fashion compatibility estimation.

\begin{table}[t]
	\begin{center}
		\begin{tabular}{|c|l|}
			\hline 
			Theme Type & Theme Tag(outfit counts)   \\
			\hline
			Occasion & 
			\makecell{Dating(4674), Travel(1706), Party(1206), \\ 
			Sports(578), School(1179), Business(2769), \\
			Home(447), Wedding(35)} \\
			\hline
			Style & 
			\makecell{Sports(491), Casual(2485), Office(1116), \\ 
			Japanese(255), US(2733), UK(1230), \\
			Girls(1994), Ladies(373), Simple(1029),\\ Nature(1765), Punk(224), Folk(37)} \\
			\hline
			Fit & 
			\makecell{Bottom(8), Small Face(17), Long Neck(30), \\ 
			White Skin(215), Thin(6184), Tall(971), \\
			Breast(30), Young(1057), Strong(6)} \\
			\hline
			Gender & 
			\makecell{Female(9502), Male(4412), Unisex(375)} \\
			\hline
		\end{tabular}
	\end{center}
	\caption{Number of outfits under the theme tags in Fashion32 dataset. 32 theme tags are grouped into 4 types including occasion, style, fit, and gender.}
	\label{tab:fashion_themes}
%	\vspace{-0.2in}
\end{table}

\section{The Fashion32 Dataset}
As aforementioned, most current fashion datasets lack the capability of theme-aware fashion compatibility learning because no fashion theme and labels of fine-grained item category are provided. Accordingly, we collected a new fashion dataset called Fashion32. There are about 13K outfits, and each of them has been labeled with multiple themes from a set of 32 themes. Also, each outfit fashion item is tagged with one of the 152 fine-grained categories. Figure~\ref{fig:FashionTheme1} shows some outfit examples from our dataset. Every single outfit has rich meta information and various labels, as well as real model pictures. To the best of our knowledge, this is the first real-world dataset carrying both theme and fine-grained category annotations for each outfit and fashion item. The annotations were labeled by fashion stylists from the brand vendors. It can be publicly accessed through the following link: http://www.larry-lai.com/fashion.html

% Data Statistics
\textbf{Dataset collection.} 
The Fashion32 dataset is crawled from the fashion channels of the e-commerce platform JD.com, one of the largest e-commerce platforms for fashion shopping. We collected 32 fashion themes as listed in Table \ref{tab:fashion_themes}. These fashion themes were proposed by fashion designers and utilized by the platform to index its products. Each collected outfit in Fashion32 is designed by fashion designers and uploaded to the platform by the brand vendors. The brand name of designers is recorded in each outfit for reference. 

We collected 13,914 outfits, as well as additional 40,667 images of fashion items and 51,415 model pictures. One outfit usually contains 2 or 3 fashion items and more than 4 pictures of a model wearing these fashion items, as shown in Figure~\ref{fig:FashionTheme1}. The model pictures are important in fashion compatibility because they demonstrate how to select fashion items to form an outfit. Each fashion item was assigned with one label for coarse-grained category (i.e., 6 categories including inner top, outer top, bottom, shoe, bag, and accessory) and one label for fine-grained category (i.e., 152 categories). This assignment was done when the vendors uploaded their outfits to the platform.  

Besides, all fashion items carry more information including product name, Stock Keeping Unit (SKU) ID, tags of design/style, tags of texture/fabric, tags of color, and a paragraph for product descriptions. To our knowledge, this dataset has the most detailed fashion labels, which can be used not only in fashion compatibility but also in the fashion image analysis. 
%Because the motivations of the fashion stylists are inducing customers to purchase their fashion items, these outfits are explicitly labeled and the quality is clean. 

\textbf{Theme tags and descriptions.} The fashion theme of an outfit can carry rich context information on it. This high-level fashion knowledge can reflect an outfit's style, occasion, or culture. Hence, in this dataset, we mainly have the following four groups of fashion themes: occasion, style, fit, and gender. There are 32 themes in total as shown in Table \ref{tab:fashion_themes}, which also lists the number of outfits collected for each theme. Each outfit is labeled with at least one theme. Also, for each outfit, we also collected a paragraph description, which explains the reason for fashion compatibility and recommendation. As an example, the Outfit C in Figure \ref{fig:FashionTheme1} is labeled with theme tags: \textit{travel, thin, nature, and female}, and its description for the recommendation looks like \textit{ ``Denim jackets open or bare shoulders, with straw hat sunglasses, full of holiday atmosphere''}. 
% 牛仔外套打开穿着或露单肩穿着，搭配草帽墨镜，洋溢十足度假氛围。
 
\textbf{Fine-grained category.} As aforementioned, the Theme-Attention is built on the fine-grained categories, as they usually carry more high-level knowledge that can be used to form a theme. Each fashion items is labeled with one of 152 fine-grained categories such as T-shirt, jacket, boots, wallet, sunglasses, and so on. The fashion designers labeled the fashion attributes for each fashion item to construct the fine-grained categories. For example, the jacket of Outfit C in Figure \ref{fig:FashionTheme1} has the fine-grained category \textit{short jacket}, and its attributes are \textit{lapel, ruffle, simple, cowboy, long sleeve, tops, female, and jacket}. Each fashion item has 7 attributes in average.

%\textbf{Difference to the existing dataset.} 

%Although the Fashion32 dataset is proposed for the research of fashion compatibility, their associated model pictures and pictures of fashion items can be used in the research of dress fitting~\cite{dress2018,hidayati2018learning} and virtual try-on~\cite{lassner2017generative,han2018viton,wang2018toward}. 

\section{Approach}
In this section, we first formulate the theme-aware compatibility learning problem. Then, we propose our framework as shown in Figure~\ref{fig:framework}, which consists of two main parts: (a) Category-Specific Embedding Network, and (b) Theme-Aware Attention Learning. 

%we aimed to predict theme-aware compatibility for fashion outfitting. For the target, we addressed our problem as .

%The goal of our method is to learn theme-aware compatibility for each of entire outfit. As shown in Figure~\ref{fig:framework}, we learn theme-aware compatibility consisting of three main parts. At the first part, we construct a common category graph based on fine-grained categories to learn outfitting knowledge from the correlations of outfit items and theme-aware associations of categories. The embedding network and masking operation encode given outfit items as non-linear embedding and project pair-wise distance to category graph edges with multiple subspace embedding correlations. 

%(a) Triplet Embedding Network is to encode each outfit as embedding representation in visual appearance space which used to calculate correlations between items of outfit for measuring compatibility (b) Shareable Embedding Masking is proposed to improve the item embedding to category-specific domain, and the shareable masking operation project category-specific embedding into multiple subspaces and reduce computation complexity (c) Theme-Aware Attentive Graph Learning was designed to learn outfit-wise compatibility with theme by assigning attentive weights on associations of category graph for each of outfit. 

%Given amount of outfits, and each of outfit composed of several items, which belongs to different fine-grained categories. 

\subsection{Problem Formulation}\label{subsec:formulation}
The proposed fashion compatibility learning framework consists of two major components: category-aware triplet embedding and attentive Theme-Attention, where category serves as a bridge connecting two components. Given a fashion outfit $O$, let $o^u$ be one of the fashion items of $O$, where the superscript denotes this item's category is $u \in C$ ( $C$ is the fine-grained category set ). Then, the theme-aware fashion compatibility of outfit $O$, given the fashion theme $P$, can be computed as, 
\begin{equation}
y_O^P = \sum d(o^u,o^v|P) \\
\: \: \:  \forall o^u,o^v \in O, u \ne v 
\end{equation}
where $u$ and $v$ are item categories, $d(o^u,o^v|P)$ is the pairwise distance between any pair of items in outfit $O$, and $d$ is computed in a category-specific embedding subspace with theme-attention. The employment of theme-attention enables our approach to obtain the outfit-wise compatibility conditional on a theme $P$, rather than simply averaging the pairwise compatibility of an outfit.

\subsection{Category-Specific Embedding Network}
%Basically, fashion compatibility can be a metric learning problem, which aim at learning a ``compatibility'' metric in an embedding space by leveraging human annotations. 
Given one pair of compatible items and one incompatible pair, a Triplet Network can learn a mapping function, which projects compatible items close to each other and incompatible ones separable in the embedding space. In general, the fine details of fashion items are important for the embedding to learn a fashion compatibility ``metric''. To better capture the details, we prefer to learn a category-specific embedding, since the compatibility measurement between one pair of categories can be different from that of another pair. In other words, a mapping function $f^{(u,v)}$ is learned to measure the compatibility between items from categories $u$ and $v$. For simplicity, in this section, the superscript $(u,v)$ will be omitted. 
 
%The goal of Projected Triplet Embedding Network is to learn the embedding of items and apply the pair-wise correlations within outfit (between items in a same outfit) to estimate compatibility. By the embedding network, outfit items will not only be encoded to get discriminate embedding in visual appearance space, but also force the correlations between items in same compatible outfit will higher than not in. Then the correlations between items willreflects the pair-wise compatibility based on visual appearance of items.
%In our task, the each outfit composed of items with different fine-grained categories, and the items of same categories also exists in different outfits. So we  create a cross-outfit embedding constraints for category graph learning. 
%Here is the detail introduction for the first part.

% cross-outfit projection constraints. 

%Suppose the outfit space is $O\sim\mathcal{O}^M$ with $M$ outfits and fine-grained category-specific embedding space is $C\sim\mathcal{C}^N$ with $N$ fine-grained categories. We denote a lots of pairs of category classes like $(u, v)$ to describe the embedding space $\mathcal{C}^{(u,v)}$. So we designed space projection $f: \mathcal{O}^M \rightarrow \mathcal{C}^{(u,v)}$ to learn fine-grained category-specific embedding space from outfit graph space. The outfit was build on outfitting items, but the category-specific embedding space describes the relevance of category-level outfit. 

Given an outfit $O_i$ and two items $o_i^u$, $o_i^v$ in $O_i$, where $u,v$ indicates the corresponding item's category, to compute the distance $d(o_i^u, o_i^v)$ in terms of compatibility, we adopt multiple layers CNN with deep residual block ~\cite{he2016deep} (see Section ~\ref{subsec:DetailsPV} for details) to embed two items into the $(u,v)$ category-specific space as $f(o_i^u;\theta)$ and $f(o_i^v;\theta)$, where $\theta$ represents the CNN parameters. As a result, the distance between two outfit items can be computed as,
\begin{equation}
d(o_i^u, o_i^v) = ||f(o_i^u;\theta) - f(o_i^v;\theta)||_2^2
\label{eq:distanceOf2} 
\end{equation}
where $d(o_i^u, o_i^v)$ is the Euclidean Distance~\cite{euclidean1980}. In the experiments (Section~\ref{subsec:DetailsPV}), we will introduce the details of different deep networks to construct non-linear projections.

%Suppose the pair of outfit items is $o_i^u, o_i^v \in O_i$. The distance of two outfit items denotes $d(o_i^u, o_i^v)$, which describes inner compatible correlations between two items with different categories $(u,v)$. To calculate the pair-wise correlations inner outfits is non-linear mapping, we adopted multiple layers CNN~\cite{he2016deep} to calculate embedding vectors $f(o_i^u;\theta)$ and $f(o_i^v;\theta)$ for the two items in a same outfit $o_i$. And the distance of them can be calculated as below:
%\begin{equation}
%d(o_i^u, o_i^v) = ||f(o_i^u;\theta) - f(o_i^v;\theta)||_2^2
%\label{eq:distanceOf2} 
%\end{equation}
%where $d(o^u, o^v)$ is the Euclidean Distance~\cite{euclidean1980} between the embedding representation of fashion items $o_i^u$ and $o_i^v$. In the experiments (as shown in~\ref{subsec:DetailsPV}), we will introduce the details of different deep networks for constructing the non-linear projection. 

% other outfit ,same category 近于不同
To learn a category-specific (i.e., $(u,v)$ category-pair) mapping function $f$, we form a training set $\mathcal{T}$ consisting of a group of training triplets $\{o_i^u, o_i^v, o_j^v\}$ by selecting two items $o_i^u$ and $o_i^v$ from outfit $O_i$ and the third item $o_j^v$ from another outfit $O_j$. The selected items are from either category $u$ or $v$. One assumption in our triplet formulation is that items $o_i^u$ and $o_i^v$ from the same outfit are compatible, while $o_j^v$ from another outfit is incompatible to the other two items. If $o_i^u$ is the anchor of a specific outfit, the optimization goal is to force the distance between items in the same outfit $d(o_i^u, o_i^v)$ closer than that of items $(o_i^u, o_j^v)$ from different outfits. Therefore, our goal during the triplet network learning is to minimize the following loss function over the training set $\mathcal{T}$: 
\begin{equation}
%\mathcal L(v, u|\theta) = 
\begin{aligned}
\mathcal Loss(o_i^u, o_i^v, o_j^v)= & \sum_u \sum_k max\{0, d(o_i^u, o_i^v) - d(o_i^u, o_j^v) + \mu\} \\
%& + \sum_v \sum_k max(0, m - d(v, u) + d(v, u_k))
\end{aligned}
\end{equation}
where $\mu$ is some margin. 
%Based on the cross-outfit loss constraint, we will finish the first graph projection that the model learn the category-specific embedding space for category graph from outfit graph.   

% \begin{figure}[t]
% 	\begin{center}
% 		%\fbox{\rule{0pt}{2in} \rule{0.9\linewidth}{0pt}}
% 		\includegraphics[width=0.95\linewidth]{Shared_Mask.pdf}
% 	\end{center}
% 	\caption{Share the embedding masks among categories.}
% 	\label{fig:Shared_Mask}
% 	\vspace{-0.2in}
% \end{figure}

\subsection{Embedding Mask}\label{subsec:SharedMask}
The category-specific embedding network attempts to learn an independent mapping function $f^{(u,v)}$ for any pair of categories $(u,v)$. It results in $|C|\times(|C|-1)$ ($C$ is the category set) number of CNN networks or embedding spaces. These individual embedding processes are less efficient because the CNNs could be highly redundant since the difference between two category-specific embeddings may be small. Therefore, instead of learning individual spaces, we propose to learn category-specific sub-spaces, which enables to learn a shared mapping function for all category-specific embedding. To this end, we further introduce a category-specific mask $m^{(u,v)}$ into the triplet embedding process. The mask serves as a gate function by selecting relevant bins to project an item to its category-specific subspace, which is depicted as $f(;\theta) \odot m^{(u,v)}$.     

Then, the distance between two items in Equation~(\ref{eq:distanceOf2}) can be represented with category-specific compatibility:
\begin{equation}
d(o_i^u, o_i^v, m^{(u,v)}) = ||f(o_i^u;\theta) \odot m^{(u,v)} - f(o_i^v;\theta) \odot m^{(u,v)}||_2^2
\label{eq:dist-sub}
\end{equation}
where $m^{(u,v)}$ is a $1 \times n$ vector, and $n$ is also the output size of feature extractor $f(;\theta)$. 

Therefore, the modified conditional triplet loss is represented as:
\begin{equation}
\begin{split}\textbf{}
%\mathcal L(v, u|\theta) = 
\begin{aligned}
& \mathcal Loss(o_i^u, o_i^v, o_j^v, m^{(u,v)}; \theta)= \\
& \sum_u \sum_k max\{0, d(o_i^u, o_i^v, m^{(u,v)}) - d(o_i^u, o_j^v, m^{(u,v)}) + \mu\} 
\end{aligned}
\end{split}
\label{eq:loss_y}
\end{equation}
The loss can be minimized by learning the embedding mask $m^{(u,v)}$ to each category pair.

\subsection{Theme-Aware Attention Learning}

Given an outfit $O$, we can compute its outfit-wise compatibility by evaluating the pairwise compatibility of all pairs items $(o^u,o^v)$ in outfit $O$. One straightforward solution is to average all pairwise compatibility as following,
\begin{equation} \label{eq:compatibility_y}
\begin{aligned}
y =  \sum d(o^u, o^v, m^{(u,v)}) / k
\end{aligned}
\end{equation}
where $o^u$ and $o^v $ are items in outfit $O$, and $k$ is total number of item pairs in $O$. This solution treats each pair equally without considering the outfit's theme tags. As shown in Figure~\ref{fig:task}, outfit A may be highly compatible without taking into account the fashion theme, while it may not be suitable for a business purpose. Therefore, we shall take into account fashion themes when measuring outfit-wise compatibility. To this end, an attentive Theme-Attention is proposed in this work. 

In fact, the theme-aware fashion compatibility is an attention problem. The theme-attention is built to link themes to pairwise compatibility and eventually enables to add theme attention to the estimation of outfit-wise compatibility. As shown in Figure~\ref{fig:framework} (c), the pairwise compatibility (e.g., node $d(o^u,o^v)$ and $d(o^u,o^t)$) is computed based on category-specific embedding. The yellow edges imply the likelihood of associating a theme (e.g., $p$ and $q$) to pairwise categories. In fact, the association likelihood is the theme-attention (e.g., $w^{(u,v)}_{P}$ ) to pairwise compatibility when aggregating all pairwise ones into the outfit-wise compatibility. Consequently, learning theme-attention is to learn the theme-attention values like $w^{(u,v)}_{P}$. 

%Given a specific theme, the attention network affects the compatibility score through its learned attention to pairwise categories as shown in Figure~\ref{fig:framework}. 
Therefore, given a fashion theme $P$, the theme-aware compatibility for an outfit $O$ can be computed by,
\begin{equation}\label{eq:compatibility_y_w}
\begin{aligned}
y^P = \sum w^{(u,v)}_{P} \cdot d(o^u, o^v, m^{(u,v)}) \\
\end{aligned}
\end{equation}
where $w^{(u,v)}_{P}$ indicates the attention weight for the category pair $u$ and $v$. Putting all $w$ together, we obtain an attention matrix $W_P$ for a given theme $P$. As we can see, $d(o^u, o^v, m^{(u,v)})$ directly measures the compatibility based on the items' appearance. Basically, it can be treated as instance-level compatibility conditional on their category. While attention matrix $W$ carries high-level human knowledge when measuring if an outfit compatible or not.   

To learn the attention network $W_P$ for theme $P$, we treat all compatible outfits associated with theme $P$ as positive examples. To obtain the negative examples, we select outfits from other themes, as well as creating an incompatible outfit by replacing one or several items in a compatible one. We treat the compatibility prediction as a classification problem, and formulate the loss function as a Cross-Entropy Loss~\cite{Goodfellow2016} as,
\begin{equation}  \label{eq:loss_y_w}
\mathcal{L}_i^P = y_i^P \cdot \log x_i^P
+ (1 - y_i^P) \cdot \log (1 - x_i^P)
\end{equation}
where $y_i^P$ is the output, and the $x_i^P$ is ground-truth of theme-aware compatibility under theme $P$ condition.

\section{Experiments}
%%%%%%%%%%%%%%%%%%%%%%%%%%%%%%%%%%%%%%%%%%%%%%%%%%%%%%%%%%%%%%%%%%%%%%
%The FashionTheme dataset provides a validation opportunity to evaluate the entire approach of Theme-Attention method. For compatibility prediction and FITB tasks, we have quantitative evaluation on different settings of our model. Moreover, we also designed subjective comparison experiment for fashion outfitting.

To evaluate the effectiveness of theme tags, we conduct experiments to compare proposed theme-aware approach and the state-of-the-art methods as the baselines. To show the generality of our method, we perform experiments not only on proposed dataset Fashion32 but also on theme-ignored dataset Polyvore\cite{han2017learning}, which has been used by several previous works.

\subsection{Implementation Details} \label{subsec:DetailsPV}
\textbf{Training} In all experiments, we use ImageNet~\cite{russakovsky2015imagenet} pre-trained ResNet-50 model~\cite{he2016deep} with the bottleneck network as the backbone model. The input image is resized to $224 \times 224$ and the output of embedding is a 1000-dimension feature vector. All models are trained on a single Tesla V100 GPU, and each input mini-batch has 32 outfits. It takes about 6 hours for 50 epochs of training. The learning rate is $1e^{-2}$ and exponentially decayed by a factor of 0.2 every 10 epochs. The optimizing strategy is SGD with a momentum of 0.9. Only model parameters with the best performance on the validation set will be saved. In Fashion32, We split the outfits into three parts: training (11,040), validation (853), and testing (2,021) sets. The experiment settings for the Polyvore dataset are the same as that for Fashion32.  The Polyvore dataset is also available to download via the link \textit{http://www.larry-lai.com/fashion.html}.

\noindent \textbf{Negative Sample} All labeled outfits in a dataset are naturally positive samples (i.e., compatible outfits). So there are no annotated negative outfits (i.e., incompatible outfits), because in real life people won't compose an incompatible outfit. To generate a negative sample, we select a positive one and substitute one item in this outfit with an item that is randomly selected from the other outfits, which carry the same category with the one in the original outfit. For training, validation and test set, the ratio of compatible and incompatible outfits are all $1:1$. During evaluating, each model is evaluated 5 times with different negative samples.

\begin{table}[t]
	\begin{center}
		\begin{tabular}{|c|l|c|c|}
			\hline
			Methods & Type & Compat. AUC(\%) & FITB Acc(\%)  \\
			\hline
			Our baseline & - & $87.37$  & $71.43$  \\
			\hline
% 			\cline{2-4}
			\multirow{5}*{Theme-Attention} & Occasion & $\textbf{94.26}$  & $76.87$  \\
			~ & Fit & $93.89$  & $\textbf{78.85}$  \\
			~ & Style & $93.84$  & $76.69$  \\
			~ & Gender & $87.21$  & $74.53$  \\
			\hline
		\end{tabular}
	\end{center}
	\caption{The compatibility learning performance comparison of Theme-Attention (Baseline) and Theme-Attention between different theme types.}
	\label{tab:theme_comparsion}
%	\vspace{-0.2in}
\end{table}
 
\subsection{Compatibility Metrics}\label{subsec:metrics}
%\textbf{Compatibility Prediction (Compat.) AUC}
\textbf{Area Under Curve (AUC)} To evaluate the model's binary prediction of compatibility. We reported the Area Under the Receiver Operating Characteristic (ROC) curve.

\noindent \textbf{Fill-in-the-blank (FITB) Accuracy}
FITB task aims at filling the most compatible item into the blank of an outfit. Each blank has 4 options in our experiment, the accuracy can be calculated for the option selection process. Our model chooses the answer by predicting scores for 4 possible outfits only substituting the blank item with different options.
 
% \paragraph{The Fashion32 Dataset}\label{p:exp_FashionTheme_dataset}
% We split the outfits in this dataset into three parts: training (11,040), validation (853), and testing (2,021). 
% The experiments are conducted on the 7 different fashion themes, including dating, travel, party, sports, school, business, and home. 
% The statistics of training, validation, and testing dataset under each theme are shown in Table~\ref{tab:theme_comparsion}

% \paragraph{Cleanup for the Polyvore Dataset}\label{Polyvore_cleanup}
% Although the previous works only experimented on 5 coarse-grained categories of the Polyvore dataset, we found that each item in the Polyvore Dataset was labeled with one of 149 fine-grained categories. 
% Furthermore, we filtered the non-wearable fashion items in the outfits (e.g. paintings or cups), since they are irrelevant to outfitting. 
% For multiple items of the same category in an outfit (e.g. multiple shoes in one outfit), only the first one will be taken. Finally, outfits with less than 3 fashion items are removed, which ensure each outfit in the dataset has at least 3 items and up to 5 items. Graph segmentation is used to split the dataset into train, validation, and test dataset. The statistics of the cleaned dataset are shown in Table~\ref{tab:PolyvoreStats}.  
% We release the cleanup version of Polyvore dataset via the link \footnote{http://www.larrylai.org/fashion.html}. 

     \begin{figure}[t] 
	\begin{center}
		\includegraphics[width=0.6\linewidth]{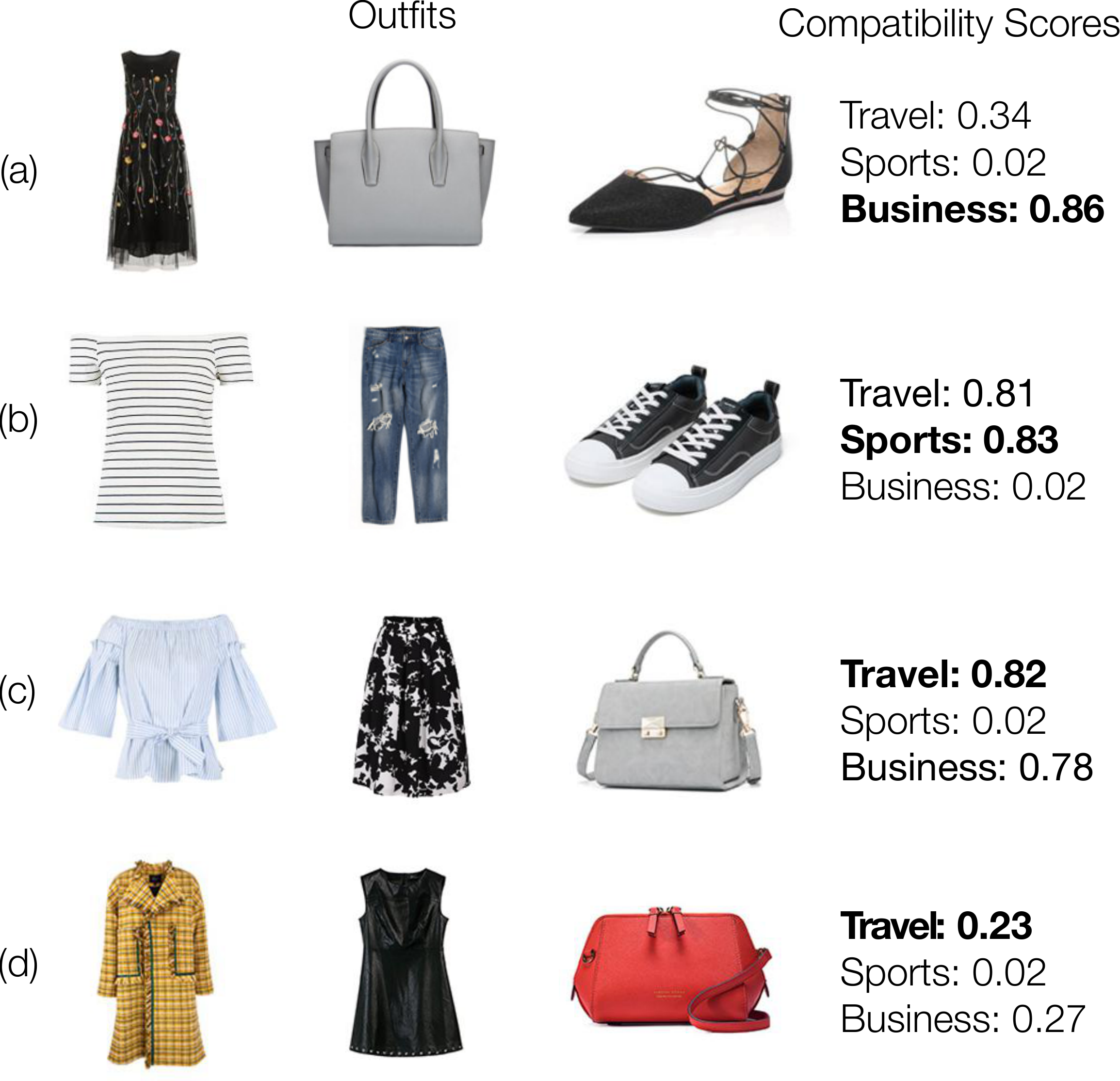} 
	\end{center}
	\caption{Some visual results of our theme-aware fashion compatibility learning. Three compatibility scores are computed given three specific themes.}
	\label{fig:themeawareScore}
	\vspace{-0.1in}
\end{figure}

% \subsection{Baselines}\label{subsec:baselines}
% \textbf{BiLSTM and VSE: } It feeds multiple CNN features of fashion items into a Bidirectional LSTM. The outputs at different time steps are taken as the prediction. The visual semantic embedding (VSE) learns a common representation between both visual and semantic information to handle the input from different sources.

% \noindent\textbf{Concatenation: } It concatenates multiple fashion features into a long vector and use two fully-connection layers to compute the compatibility score.

% \noindent\textbf{Pooing: } It uses average pooling to aggregate multiple fashion features into a single vector, then use two fully-connection layers to compute the compatibility score.

% \noindent\textbf{Type-Aware: } It learns category-specific embedding in Equation~(\ref{eq:loss_y}) but ignores the theme-aware attention $w^{(u,v)}_{P}$.

% \noindent\textbf{Theme-Attention: } There are \textbf{baseline} version and different group version (\textbf{occasion}, \textbf{style}, \textbf{fit}, \textbf{gender}) of Theme-Attention. In baseline version, the theme-aware attention weight of a single pair of category is the same for all themes. In other versions, the attention weights are different according to different themes.

\subsection{Quantitative Experiments on Fashion32 Dataset}
%For the settings of shared masks in Section~\ref{subsec:SharedMask}, we clustered the 152 fine-grained categories into 6 coarse-grained categories including inner top, outer top, bottom, shoe, bag, and accessory, see Table~\ref{tab:fineGrainedJD} in the appendix. 
%A 6x6 coarse-grained projection matrices is equal to 36 shared masks. 
%Because there are 152 fine-grained categories, the size of $w^{(u,v)}_{P}$ is $152 \times 152$. 

To verify the effectiveness of our theme-aware fashion compatibility model, we also implemented a \textbf{Baseline} version of the \textbf{Theme-Attention} method. Both of them minimize the compatibility loss via Equation~(\ref{eq:loss_y_w}) and Equation~(\ref{eq:loss_y}), respectively. The Theme-Attention is built on the Baseline with additional theme-aware attention.

Table~\ref{tab:theme_comparsion} shows the AUC and FITB scores of both methods. In terms of AUC scores, theme-attention method achieved better performance than our Baseline in almost all groups of themes except the gender group.
% Note that we run every experiment five times, and the average score is followed by the standard deviation.
In general, FITB scores are proportional to AUC scores. Especially, theme-Attention achieves 6.89\% of AUC increase for the occasion theme group. The results successfully demonstrate that theme-attention method is able to improve the quality of fashion compatibility. In addition, from the results, we can observe that theme-attention can perform better on some distinctive themes. For example, outfits of ``sports'' usually consist of T-shirt, shorts, or running shoes, which make the We theme-aware weights easy to be learned. As a comparison, the Gender theme group does not help fashion compatibility estimation. We conjecture that ``Female'' and ``Male'' are the distinctive themes because it is easy to tell if an outfit designed for male or female based on its visual appearance. Adding this incapable attention to the model actually hurt the performance. That is why its performance is worse than our Baseline.      

In terms of the FITB scores, the Theme-Attention method can achieve up to 7.42\% improvement compared to the Baseline. Overall, the Theme-Attention method can effectively learn the theme-aware fashion compatibility under a theme with narrow variations of category combinations. 

% Table~\ref{tab:theme_comparsion} compares the AUC and FITB performance of theme-aware or theme-ignored methods. Specifically, we show the results of proposed Theme-Attention with 4 runs on different types of theme tags individually: occasion, style, fit and gender. In general, it improve the performance comparing to the method without theme-aware attention. In terms of the AUC scores, the Theme-Attention method achieved a better performance than the Type-Aware method with [XX] averaged improvement over occasion, fit, style situation. However it performs inferior with gender theme tags. The reason is the cooccurence of fine-grained category from different genders has limited rule for learning compatibility, especially for female outfits. In terms of FITB scores, it performs proportionally to the AUC scores.
% \noindent\textbf{Theme tags affect the compatibility.} 

Figure~\ref{fig:themeawareScore} illustrates some visual examples of our theme-aware fashion compatibility learning results. As we can see, our model generates different compatibility scores for an outfit given different themes. The theme with the highest score implies it is the most relevant theme to this outfit. Hence, outfit (a), (b), and (c) are correctly detected as business, sports, and travel, respectively. Example (d) is a negative example without a theme. Also, the result of (b) tells us an outfit can be suitable for multiple themes, i.e., both travel and sports are good for (b). This visualization further verifies the proposed theme-aware fashion compatibility.

\begin{table*}[t]
	\begin{center}
		\begin{tabular}{|l|c|c|c|c|} 
			\hline
			      & \multicolumn{2}{c|}{\textbf{Fine-grained Polyvore}}  & \multicolumn{2}{c|}{\textbf{Fashion32}} \\ 
			\hline
			Method & Compat. AUC(\%) & FITB Acc(\%) & Compat. AUC(\%) & FITB Acc(\%) \\
			\hline
			BiLSTM \cite{han2017learning} & $74.44 \pm 0.95$ & $45.41 \pm 0.40$ & $85.12 \pm 0.26$ & $60.14 \pm 0.97$ \\
			BiLSTM+VSE \cite{han2017learning}  & $74.82 \pm 0.63 $ & $46.02 \pm 0.62$ & - & - \\
			Concatenation \cite{Tangseng2018} & $83.40  \pm 0.48$ & $52.91 \pm 0.59$ & $93.82 \pm 0.34$ & $ 69.23 \pm 1.42 $ \\
			Pooling \cite{li2017mining} & $\textbf{88.35}  \pm 0.26$ & $57.28 \pm 0.313$ & $92.97 \pm 0.43$ & $ 69.28 \pm 0.70 $ \\
			Type-Aware \cite{Vasileva2018} & $84.90 \pm 0.52$ & $57.06 \pm 1.70 $ & $88.42 \pm 0.47$ & $71.15 \pm 1.90 $ \\
			
			%Category-Graph (15-mask) & $85.51 \pm 0.34$ & $59.11 \pm 0.79$ \\
			Our Baseline & $85.85 \pm 0.37$ & $\textbf{60.66} \pm 0.84$ & $87.37 \pm 0.52$ & $71.43 \pm 0.36$ \\
			Theme Attention & - & - & $\textbf{94.26} \pm 0.62$ & $\textbf{76.87} \pm 0.87$ \\
			\hline
		\end{tabular}
	\end{center}
	\caption{The comparison between different methods on Polyvore and Fashion32 dataset.}
	\label{tab:comparsionPV}
	\vspace{-0.1in}
\end{table*}

\subsection{Subjective Experiment on Online Fashion Shop}
We further evaluate our model's performance by calculating the click rates of customers' browsing history, which is widely used by many e-commence platforms. An outfit with a higher click rate will have higher compatibility score. Unlike previous experiments, this one is more subjective.

\begin{figure}[t]
	\begin{center}
		\includegraphics[width=0.6\linewidth]{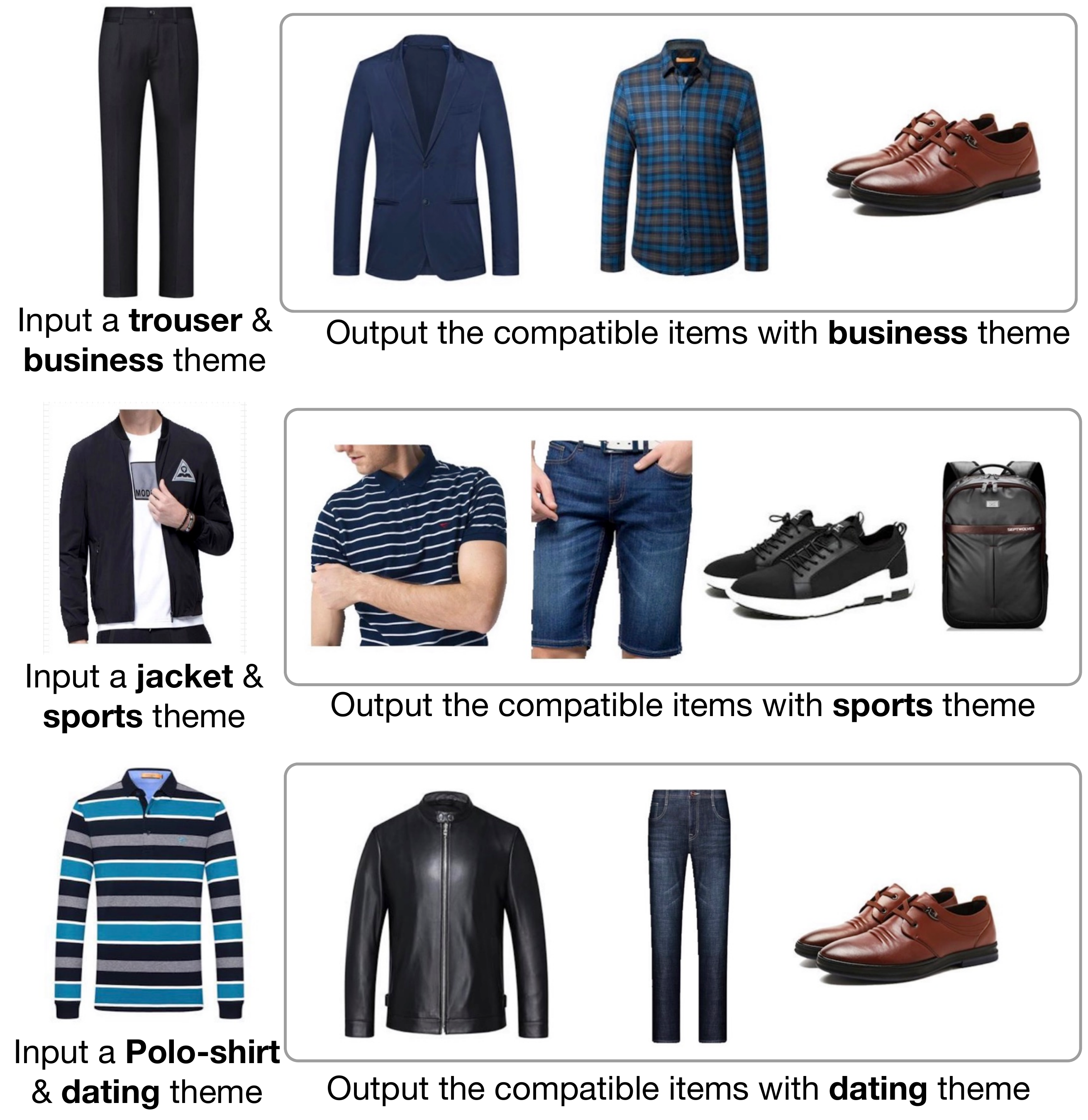} 
	\end{center}
	\caption{Given an item and a theme, the Theme-Attention method searches on the pool of online shop to recommend the compatible outfits.}
	\label{fig:subjective}
	\vspace{-0.1in}
\end{figure}

We collected 500 fashion items from an online fashion shop %(i.e., http://www.XXXX.com) 
as the searching pool. Given a fashion item and a target theme, two experimental algorithms (i.e., our theme-attention and non-theme method) will recommend 3 to 5 items from the search pool to generate a complete outfit. Figure~\ref{fig:subjective} shows some recommended outfits. As we see, the results are not only visually compatible but also suitable for the given theme. During the evaluation, 5 subjects are assigned to conduct the click rate experiments. The subjects were asked to tell which is more compatible given a fashion theme. To avoid subjects' bias on one recommendation algorithm, we randomly switched the order of two recommendations. Each subject evaluated 200 outfits on three themes: business, travel, and sports. 

The final results show that Theme-Attention method is better than non-theme method with 8.6\% improvement in terms of click rates. It further demonstrates that our approach can recommend theme-compatible outfits from a pool of new fashion items which have not to be seen during the training.

% . And in real online scenarios or life, fashion outfitting usually followed outfitting rules or suggestions by designers or outfitters. So we designed subjective evaluation experiment, to compare our proposed approach and collected outfitting rules generated by expert designers from JD.com. As shown in Figure~\ref{fig:subjective}, a user will see the two outfits and rating them at same time. The outfits were extracted from the positive examples of FashionTheme dataset . The comparison between rule-based method and proposed graph modeling approach designed for validating 200 outfits over 5 subjects. Meanwhile, the evaluation let test persons score the right one or the left one is better with considering the ``Condition'' of outfitting. By human observation and comparison, our method is better than the rule-based 8.6\% in terms of Click Rates. 
     
%%%%%%%%%%%%%%%%%%%%%%%%%%%%%%%%%%%%%%%%%%%%%%%
\subsection{Performance Comparison}
Since we are the first to work on the theme-aware fashion compatibility problem, it is very difficult to directly compare our approach with previous work. Since our Baseline is non-theme version of our approach, we can apply our Baseline to Polyvore dataset for comparison.  However, our Baseline requires fine-grained categories which Polyvore does not have. To this end, we improved Polyvore dataset to fine-grained Polyvore ( please refer to our supplemental material for this improved version ). In addition, we also managed to run previous approaches on our Fashion32 dataset without leveraging the theme information.  

Table~\ref{tab:comparsionPV} illustrates the detailed performance comparison on both fine-grained Polyvore and Fashion32 dataset. On fine-grained Polyvore dataset, as we can see, our Baseline is comparable to previous approaches for both compatibility prediction and FITB tasks. As our Baseline is actually a non-theme version of our approach, this comparison is somewhat meaningful. 
On Fashion32 dataset, our theme-attention outperforms the state-of-the-arts approaches. In terms of FITB, our approach is 7\% better than state-of-the-art approaches. Although this is not a direct comparison, the results still demonstrate the advantages of theme-aware fashion compatibility. Previous approaches do not have a mechanism to leverage theme information for fashion compatibility.

\section{Conclusions}
To solve the theme-aware fashion compatibility problem, in this paper, we collected the first theme-matters fashion dataset, which contains 13K outfits in total over 32 themes and 152 fine-grained category classes. We further propose a novel benchmark, which leverages the attentive Theme-Attention built on category-specific embedding, to learn theme-aware fashion compatibility. we evaluate our approaches by both objective and subjective experiments. Compared with the baseline, our method Theme-Attention achieved 94.26\% AUC in outfit compatibility prediction and 78.85\% accuracy in FITB task, respectively. Comparing to several recent works on the Polyvore dataset, the Baseline version of the Theme-Attention method also achieves competitive on compatibility prediction and FITB tasks.

%In future work, we would explore more studies on interpretive modeling for fashion outfitting. Because human will easily to understand the outfitting results while more explanatory techniques will be incorporated into fashion presentation and learning. 

\newpage

\section{Supplementary Materials}
\subsection{A. Detailed Comparison between theme attention learning and non-theme method}
Table.\ref{tab:occasion} to Table.~\ref{tab:gender} provide detailed results for each theme tag group. Not only the overall performance but also the results for each kind of theme tags (e.g. sports, dating, etc.) are presented. It can be seen that theme attention method show superiority in almost every theme tags.

\subsection{B. The statistics of Fine-grained Polyvore}
We cleaned up Polyvore dataset with fine-grained categories. Polyvore dataset does not have theme tags, but has 149 fine-grained categories to satisfy the baseline version of the Theme-Attention method. Firstly, we filtered the 692 non-wearable fashion items in the outfits (e.g. paintings or cups). For multiple items of the same category in an outfit (e.g. multiple shoes in one outfit), only the first one will be taken. Finally, outfits with less than 3 fashion items are removed. The cleaned dataset will be publically available at www.larry-lai.com/fashion.html, its statistics are shown in Table.~\ref{tab:PolyvoreStats}.

\subsection{C. The Panel Designed for Click Rate Experiment}

\begin{figure}[!htb]
	\begin{center}
		\includegraphics[width=0.9\linewidth]{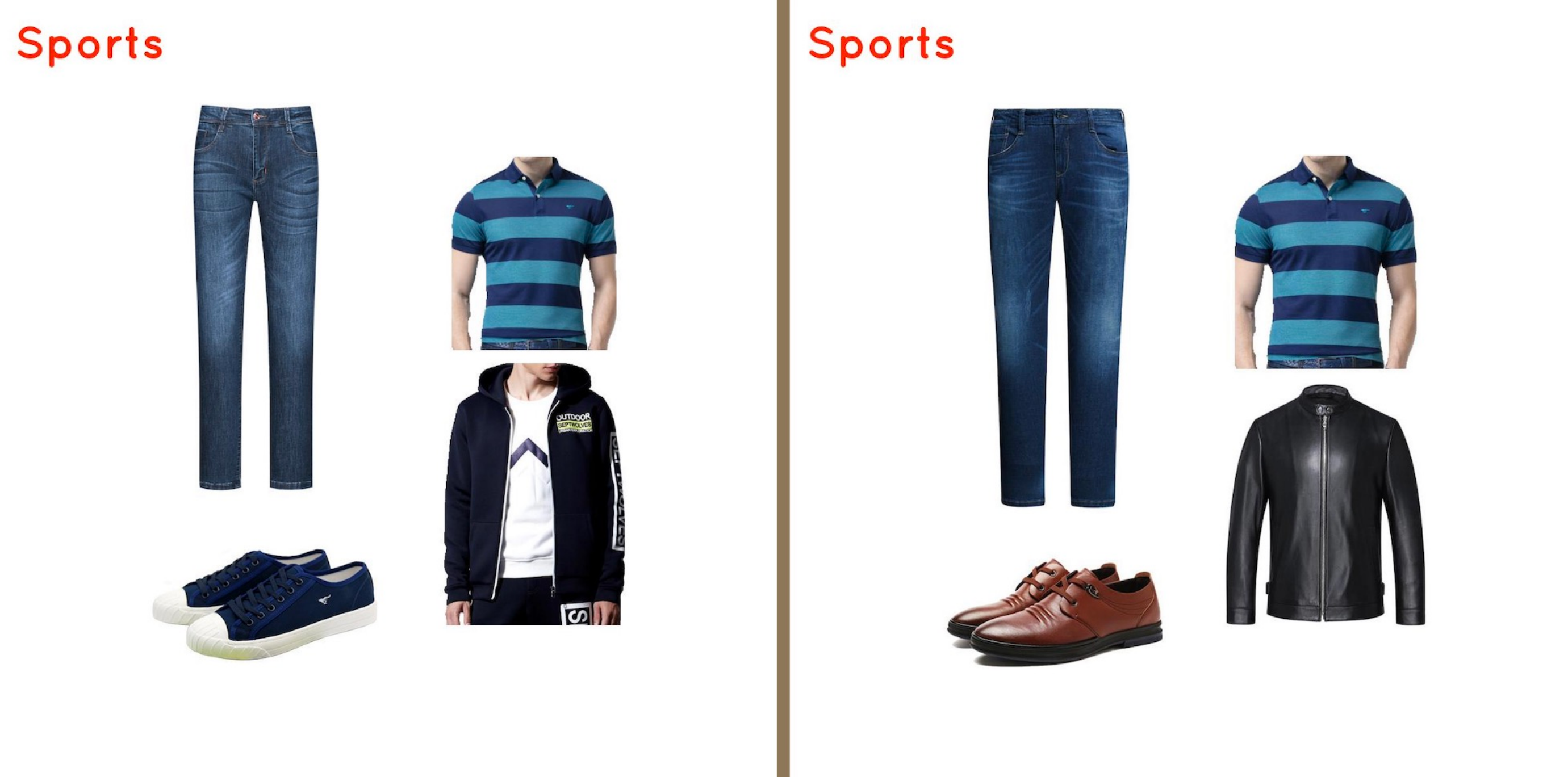}
	\end{center}
	\caption{The panel for subjects to click on which outfit is more compatible with considering the sports theme.}
	\label{fig:side-by-side}
	%\vspace{-0.2in}
\end{figure}

In the click rate experiment, the subjects were asked to click on the outfit, which is more compatible given a fashion theme, given two outfits as shown in Figure.~\ref{fig:side-by-side}. To avoid subjects to have any bias on one side, we randomly switched the order of two recommendations.

\subsection{D. Visual results for outfit compatibility with respect to different themes}
\begin{figure}[!htb] 
	\begin{center}
		\includegraphics[width=0.6\linewidth]{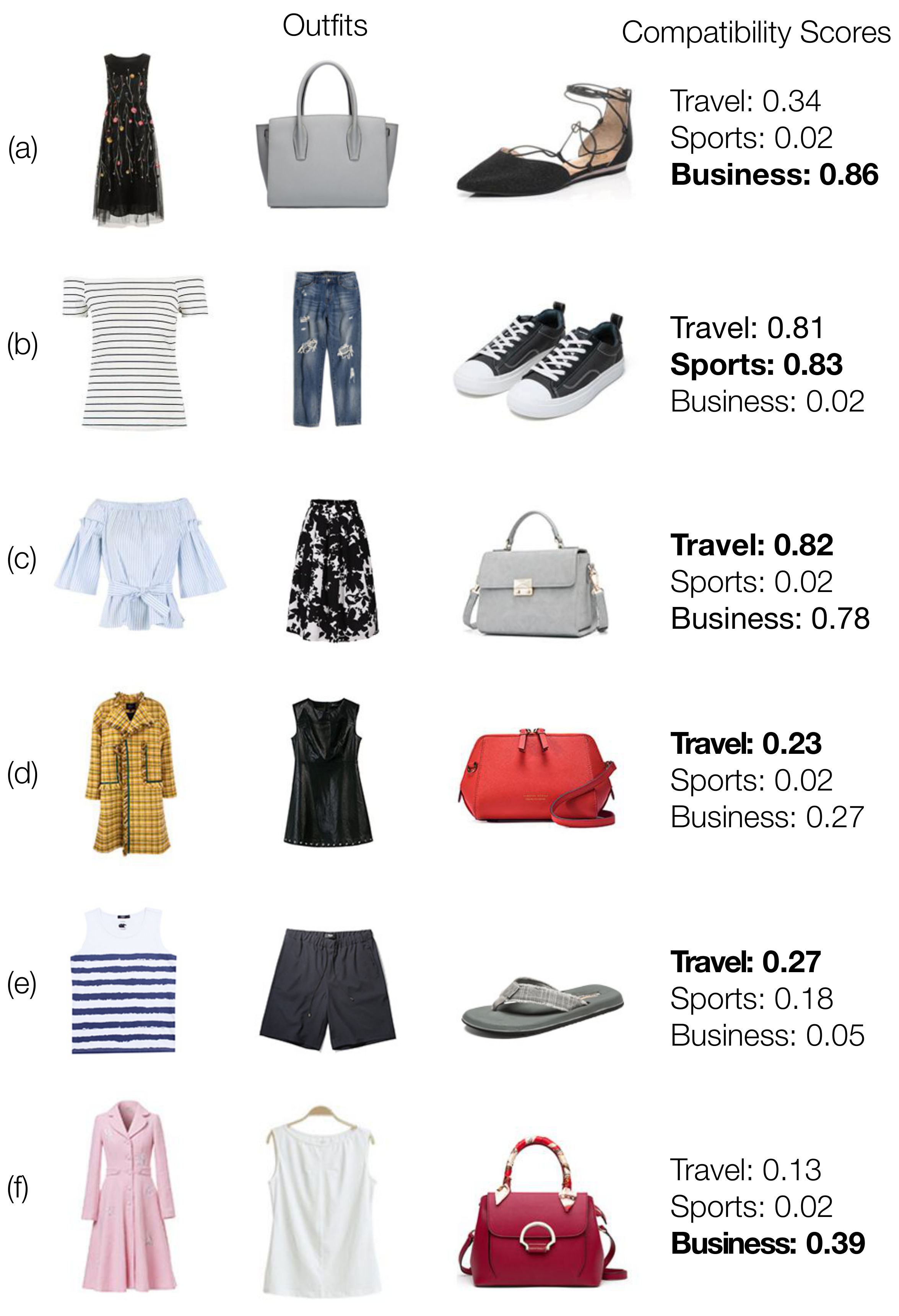} 
	\end{center}
	\caption{Some visual results of our theme-aware fashion compatibility learning. Three compatibility scores are computed given three specific themes.}
	\label{fig:themeawareScore}
	\vspace{-0.1in}
\end{figure}

\begin{table*}[ht]
	\begin{center}
		\begin{tabular}{|l|c|c|c|c|c|c|c|}
			\hline
			% \multirow{1}{*}{ \textbf{Fashion}}
			Fashion
			& \multicolumn{3}{c|}{ \textbf{Outfits}} & \multicolumn{2}{c|}{ \textbf{Type-Aware} } & \multicolumn{2}{c|}{ \textbf{Theme-Graph} }  \\ 
			\cline{2-8}
			%  \multirow{1}{*}{ \textbf{Themes}} 
			Themes
			& Train & Validation & Test & Compat. AUC(\%) & FITB Acc(\%) & Compat. AUC(\%) & FITB Acc(\%)  \\
			\hline
			Dating & 3744 & 269 & 661 & $88.05 \pm 0.79$  & $72.62 \pm 1.44$ & $\textbf{93.29} \pm 0.57$  & $\textbf{74.36} \pm 0.67$  \\
			%\hline
			Travel & 1351 & 120 & 235 & $90.69 \pm 1.11$  & $76.75 \pm 1.74$ & $\textbf{94.82} \pm 0.75$  & $\textbf{79.39} \pm 2.16$  \\
			%\hline
			Party & 940 & 88 & 178 & $89.28 \pm 3.34$  & $73.64 \pm 1.45$ & $\textbf{92.54} \pm 1.29$  & $\textbf{74.78} \pm 3.63$  \\
			%\hline
			Sports & 470 & 30 & 78 & $78.28 \pm 3.01$  & $68.60 \pm 6.13$ & $\textbf{98.61} \pm 0.52$  & $\textbf{80.04} \pm 5.42$  \\
			%\hline
			School & 938 & 86 & 155 & $88.13 \pm 2.25$  & $74.42 \pm 3.57$ & $\textbf{96.92} \pm 1.15$  & $\textbf{82.52} \pm 1.03$  \\
			%\hline
			Business & 2197 & 167 & 405 & $90.74 \pm 1.02$  & $72.07 \pm 1.01$ & $\textbf{94.75} \pm 0.95$  & $\textbf{78.18} \pm 2.00$  \\
			%\hline
			Home & 355 & 37 & 55 & $85.86 \pm 4.73$    & $\textbf{78.63} \pm 8.60$ & $\textbf{91.60} \pm 4.76$  & $73.16 \pm 4.36$  \\
			%Wedding & 51 & 5 & 6 & -  & - & - & - \\
			\hline
			\textbf{Total} & 9995 & 797 & 1767 & $88.41 \pm 0.47$  & $71.15 \pm 1.90$ & $\textbf{94.26} \pm 0.62 $  & $\textbf{76.87} \pm 0.87$  \\
			\hline
		\end{tabular}
	\end{center}
	\caption{The performance comparison over occasion theme group.}
	\label{tab:occasion}
\end{table*}

\begin{table*}[t]
	\begin{center}
		\begin{tabular}{|l|c|c|c|c|c|c|c|}
			\hline
			% \multirow{1}{*}{ \textbf{Fashion}}
			Fashion
			& \multicolumn{3}{c|}{ \textbf{Outfits}} & \multicolumn{2}{c|}{ \textbf{Type-Aware} } & \multicolumn{2}{c|}{ \textbf{Theme-Graph} }  \\ 
			\cline{2-8}
			%  \multirow{1}{*}{ \textbf{Themes}} 
			Themes
			& Train & Validation & Test & Compat. AUC(\%) & FITB Acc(\%) & Compat. AUC(\%) & FITB Acc(\%)  \\
			\hline
			Bottom & 7 & 0 & 1 & -  & - & -  & - \\
			%\hline
			Small Face & 10 & 1 & 6 & $74.17 \pm 17.56$  & $49.67 \pm 26.47$ & $\textbf{82.50} \pm 18.71$  & $\textbf{56.67} \pm 34.90$  \\
			%\hline
			Long Neck & 27 & 0 & 3 & $40.00 \pm 20.00$  & $0.00  \pm 0.00$ & $\textbf{80.00} \pm 24.49$  & $\textbf{86.67} \pm 16.33$  \\
			%\hline
			White Skin & 177 & 9 & 29 & $44.30 \pm 7.42$  & $63.65 \pm 11.99$ & $\textbf{84.59} \pm 5.37$  & $\textbf{73.72} \pm 3.60$  \\
			%\hline
			Thin & 4938 & 401 & 845 & $90.75 \pm 1.07$  & $76.24 \pm 1.18$ & $\textbf{94.63} \pm 0.56$  & $\textbf{79.70} \pm 0.96$  \\
			%\hline
			Tall & 771 & 58 & 142 & $90.54 \pm 2.14$  & $70.38 \pm 3.55$ & $\textbf{93.94} \pm 0.88$  & $\textbf{75.65} \pm 3.03$  \\
			%\hline
			Breast & 26 & 3 & 1 & -    & - & -  & -  \\
			Young & 846 & 72 & 139 & $86.62 \pm 3.83$    & $76.76 \pm 2.28$ & $\textbf{93.56} \pm 1.61$  & $\textbf{78.95} \pm 1.33$  \\
			Strong & 4 & 0 & 2 & -  & - & -  & -  \\
			\hline
			\textbf{Total} & 6806 & 544 & 1168 & $89.67 \pm 0.90$  & $74.17 \pm 1.26$ & $\textbf{93.89} \pm 0.55$  & $\textbf{78.85} \pm 1.01$  \\
			\hline
		\end{tabular}
	\end{center}
	\caption{The performance comparison over fit theme group.}
	\label{tab:fit}
\end{table*}

\begin{table*}[ht]
	\begin{center}
		\begin{tabular}{|l|c|c|c|c|c|c|c|}
			\hline
			% \multirow{1}{*}{ \textbf{Fashion}}
			Fashion
			& \multicolumn{3}{c|}{ \textbf{Outfits}} & \multicolumn{2}{c|}{ \textbf{Type-Aware} } & \multicolumn{2}{c|}{ \textbf{Theme-Graph} }  \\ 
			\cline{2-8}
			%  \multirow{1}{*}{ \textbf{Themes}} 
			Themes
			& Train & Validation & Test & Compat. AUC(\%) & FITB Acc(\%) & Compat. AUC(\%) & FITB Acc(\%)  \\
			\hline
			Sports & 406 & 7 & 78 & $89.40 \pm 2.04$  & $77.70 \pm 6.53$ & $\textbf{96.46} \pm 1.24$  & $\textbf{78.69} \pm 4.34$  \\
			%\hline
			Casual & 1945 & 184 & 356 & $87.78 \pm 1.57$  & $76.54 \pm 0.65$ & $\textbf{93.60} \pm 0.71$  & $\textbf{76.52} \pm 1.80$  \\
			%\hline
			Office & 908 & 64 & 144 & $94.06 \pm 1.30$  & $76.70 \pm 4.53$ & $\textbf{95.24} \pm 1.85$  & $\textbf{81.44} \pm 0.69$  \\
			%\hline
			Japanese & 200 & 11 & 44 & $84.64 \pm 0.70$  & $73.42 \pm 4.81$ & $\textbf{90.68} \pm 1.64$  & $\textbf{74.91} \pm 4.97$  \\
			%\hline
			US & 2182 & 151 & 400 & $90.81 \pm 0.96$  & $75.58 \pm 1.00$ & $\textbf{93.80} \pm 0.97$  & $\textbf{78.65} \pm 1.73$  \\
			%\hline
			UK & 968 & 75 & 187 & $89.27 \pm 3.14$  & $74.24 \pm 0.58$ & $\textbf{94.58} \pm 1.23$  & $\textbf{77.01} \pm 1.21$  \\
			%\hline
			Girls & 1600 & 116 & 278 & $88.02 \pm 0.56$    & $75.12 \pm 1.79$ & $\textbf{94.03} \pm 0.46$  & $\textbf{76.29} \pm 1.72$  \\
			Ladies & 282 & 32 & 59 & $91.03 \pm 2.76$    & $71.55 \pm 6.96$ & $\textbf{95.27} \pm 2.24$  & $\textbf{70.85} \pm 6.07$  \\
			Simple & 822 & 63 & 144 & $92.28 \pm 0.74$    & $78.29 \pm 2.60$ & $\textbf{93.68} \pm 1.37$  & $\textbf{76.79} \pm 1.96$  \\
			Nature & 1398 & 113 & 254 & $89.43 \pm 1.67$    & $73.01 \pm 4.82$ & $\textbf{93.68} \pm 1.47$  & $\textbf{73.03} \pm 2.83$  \\
			Purk & 180 & 17 & 27 & $83.79 \pm 6.42$    & $74.05 \pm 5.56$ & $\textbf{90.81} \pm 3.21$  & $\textbf{77.62} \pm 5.64$  \\
			Folk & 29 & 3 & 5 & $68.33 \pm 18.56$    & $\textbf{87.00} \pm 16.61$ & $\textbf{83.33} \pm 21.08$  & $37.00 \pm 28.91$  \\
			\hline
			\textbf{Total} & 10920 & 836 & 1976 & $89.47 \pm 0.61$  & $75.43 \pm 0.44$ & $\textbf{93.84} \pm 0.29$  & $\textbf{76.69} \pm 0.78$  \\
			\hline
		\end{tabular}
	\end{center}
	\caption{The performance comparison over style theme group.}
	\label{tab:style}
\end{table*}

\begin{table*}[ht]
	\begin{center}
		\begin{tabular}{|l|c|c|c|c|c|c|c|}
			\hline
			% \multirow{1}{*}{ \textbf{Fashion}}
			Fashion
			& \multicolumn{3}{c|}{ \textbf{Outfits}} & \multicolumn{2}{c|}{ \textbf{Type-Aware} } & \multicolumn{2}{c|}{ \textbf{Theme-Graph} }  \\ 
			\cline{2-8}
			%  \multirow{1}{*}{ \textbf{Themes}} 
			Themes
			& Train & Validation & Test & Compat. AUC(\%) & FITB Acc(\%) & Compat. AUC(\%) & FITB Acc(\%)  \\
			\hline
			Female & 7547 & 587 & 1368 & $\textbf{86.80} \pm 0.57$  & $70.35 \pm 0.67$ & $83.51 \pm 0.95$  & $\textbf{70.81} \pm 0.71$  \\
			%\hline
			Male & 3493 & 266 & 653 & $\textbf{95.60} \pm 0.38$  & $\textbf{82.95 }\pm 1.79$ & $94.38 \pm 0.72$  & $82.35 \pm 0.83$  \\
			\hline
			\textbf{Total} & 11040 & 853 & 2021 & $\textbf{89.70} \pm 0.34$  & $74.42 \pm 0.42$ & $87.21 \pm 0.91$  & $\textbf{74.54} \pm 0.74$  \\
			\hline
		\end{tabular}
	\end{center}
	\caption{The performance comparison over gender theme group.}
	\label{tab:gender}
\end{table*}

\begin{table*}
	\centering
	\begin{tabular}{|l|l|ccccc|cc|}  
		\hline
		&       & Top & Bottom & Shoe  & Bag   & Accessory & Item  & Outfit \\
		\hline\hline
		% \multirow{3}{*}{Before} 
		Before & Train &   -   &    -   &   -    &   -    &   -    & 114806 & 17316 \\
		& Val &   -    &  -     &   -    &   -    &   -    & 9070  & 1497 \\
		& Test  &   -    &   -    &   -    &   -    &   -    & 18604 & 3076 \\
		\hline
		% \multirow{3}{*}{After} 
		After & Train & 13764 & 14849 & 15268 & 12640 & 12093 & 68614 & 16176 \\
		& Val & 962   & 1052  & 1124  & 948   & 823   & 4904  & 1196 \\
		& Test  & 2000  & 2153  & 2314  & 1994  & 1712  & 10173 & 2463 \\
		\hline
	\end{tabular}
	\centering
	\caption{The statistics of items and outfits in the Polyvore dataset before and after data cleaning. The original Polyvore dataset does not have type labeling, so its type statistics are missing.}
	\label{tab:PolyvoreStats}
\end{table*}

\clearpage

\newpage
\bibliographystyle{unsrt} 
% \bibliographystyle{arxiv} 


\bibliography{Theme-Graph}
\end{document}